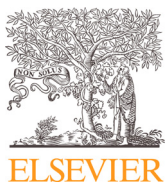
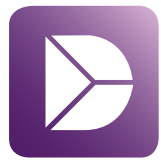

Data Article

# Performance evaluation results of evolutionary clustering algorithm star for clustering heterogeneous datasets

Bryar A. Hassan [a,c,∗], Tarik A. Rashid [b], Seyedali Mirjalili [d,e]

[a] *Kurdistan Institution for Strategic Studies and Scientific Research, Sulaimani, Iraq*
[b] *Computer Science and Engineering Department, University of Kurdistan Hewler, Erbil, Iraq*
[c] *Department of Computer Networks, Technical College of Informatics, Sulaimani Polytechnic University, Sulaimani, Iraq*
[d] *Centre for Artificial Intelligence Research and Optimisation, Torrens University, Australia*
[e] *Yonsei Frontier Lab, Yonsei University, Seoul, Korea*

a b s t r a c t

This article presents the data used to evaluate the performance of evolutionary clustering algorithm star (ECA∗) compared to five traditional and modern clustering algorithms. Two experimental methods are employed to examine the performance of ECA∗ against genetic algorithm for clustering++ (GENCLUST++), learning vector quantisation (LVQ), expectation maximisation (EM), K-means++ (KM++) and K-means (KM). These algorithms are applied to 32 heterogenous and multi-featured datasets to determine which one performs well on the three tests. For one, ther paper examines the efficiency of ECA∗ in contradiction of its corresponding algorithms using clustering evaluation measures. These validation criteria are objective function and cluster quality measures. For another, it suggests a performance rating framework to measurethe the performance sensitivity of these algorithms on varos dataset features (cluster dimensionality, number of clusters, cluster overlap, cluster shape and cluster structure). The contributions of these experiments are two-folds: (i) ECA∗ exceeds its counterpart aloriths in ability to find out the right cluster number; (ii) ECA∗ is less sensitive towards dataset features compared to its com-petitive techniques. Nonetheless, the results of the experiments performed demonstrate some limitations in the ECA*: (i) ECA* is not fully applied based on the premise that no prior knowledge exists; (ii) Adapting and utilising ECA* on several real applications has not been achieved yet.

∗ Corresponding author at: Kurdistan Institution for Strategic Studies and Scientific Research, Sulaimani, Iraq.
*E-mail address:* bryar.hassan@kissr.edu.krd (B.A. Hassan).

## Specifications Table

| | |
|---|---|
| Subject | Soft Computing |
| Specific subject area | Evolutionary clustering algorithm, machine learning, clustering in data mining |
| Type of data | Tabulted data |
| How data were acquired | Comuter machies: written Java program, software, application program |
| Data format | Raw and tabulated data |
| Parameters for data collection | • Cluster quality measures (NMI, SCI and CI) for ECA* with its competitive algorithms for the run average of 30;<br>• Objective function measures ($\varepsilon$- ratio, nMSE and SSE) for ECA* with its counterpart echniques for the run average of 30;<br>• Priary statistical measures of the 30 avrage run achived by ECA* for Execution time, InterCluster and IntraCluster (Average: mean solution, Best: the best-solution, Worst: the worst solution);<br>• Total mean performance rate of ECA* compared to its counterpart methods for 32 datasets based on five dataset properties (class overlap, number of classes, class dimensionality, class structure, and class shape). |
| Description of data collection | The data collection was from the written Java codes by the authors and Weka packages for executing ECA* on 32 heterogenous and multi-featured datasets against its counterpart algorithms. The algorithms are GENCLUST++, LVQ, EM, KM++ and KM. Each algorithm was run 30 times on 32 dataset problems to evaluate its performance. The collected data are published on Mendeley Data: A. Rashid, Tarik; Hassan, Dr Bryar (2021), "Custering Results of evolutionary clustering algorithm star for clustering heterogeneous datasets", Mendeley Data, V2, doi:10.17632/bsn4vh3zv7.2 |
| Data source location | Kurdistan Institution for Strategic Studies and Scientific Research (KISSR), Sulaimani, Kurdistan Region, Iraq, 35.521700, 45.466605 |
| Data accessibility | Raw tabulated data are accessible on Mendeley Data: A. Rashid, Tarik; Hassan, Dr Bryar (2021), "Custering Results of evolutionary clustering algorithm star for clustering heterogeneous datasets", Mendeley Data, V2, doi:10.17632/bsn4vh3zv7.2 |
| Related research article | B.A. Hassan, T.A. Rashid, A multidisciplinary ensemble algorithm for clustering heterogeneous datasets, Neural Comput. Appl. (2021). doi:10.1007/s00521-020-05649-1. |

## Value of the Data

- The validation of clustering results of ECA* and its counterparg algorithms;
- This data revealed the success of ECA* in comparison to its competitors to encourage sholars for further research on ECA appled on real-world applications.
- Two experimental methods on 32 heterogenous and multi-featured benchmarking datasets are used to examine the clustering results of ECA* compared to GENCLUST++, LVQ, EM, KM++ and KM;
- Performance evaluation, and performance rating framework of ECA* compared to its competitors are presented.

# 1. Data Description

This section presents the performance evaluation of ECA* against its competitors using cluster quality measures (NMI, CSI and CI) and objective functions measures (ratio, SSE and nMSE) on 32 varied datasets.The datasets are challenging for several clustering algorithms, yet they easy enough to determine the correct cluster centroids for a proper clustering algorithm. The benchmarking datasets are well-known data and publicly accessible in [1]. The clustering benchmark datasets used with their dimensions and data properites are presented in [2].

Futuremore, the data was collected from the written Java codes by the authors, and Weka packages for executing ECA* on 32 heterogenous and multi-featured datasets against its counterpart algorithms (GENCLUST++, LVQ, EM, KM++ and KM). Each algorithm was run 30 times on the 32 dataset problems in order to test the effiicney of ECA* with its copetitors.

- ECA Star.zip;
- KM.zip
- KM++.zip
- EM.zip
- LVQ.zip
- GENCLUST++.zip

Each of the above zip files contains 32 XLSX files of clustering results to its respective algorirhtm examined on 32 benchmark datasets. The files are: A1.xlsx, A2.xlsx, A3.xlsx, Birch1.xlsx, dim032.xlsx, dim064.xlsx, dim128.xlsx, dim256.xlsx, dim512.xlsx, dim1024.xlsx, g2-16-10.xlsx, g2-16-30.xlsx, g2-16-60.xlsx, g2-16-80.xlsx, g2-16-100.xlsx, g2-1024-10.xlsx, g2-1024-30.xlsx, g2-1024-60.xlsx, g2-1024-80.xlsx, g2-1024-100.xlsx, S1.xlsx, S2.xlsx, S3.xlsx, S4.xlsx, Aggregation.xlsx, Compound.xlsx, D21.xlsx, Flame.xlsx, Jain.xlsx, Pathbased.xlsx, R15.xlsx, and Unbalance.xlsx

The evaluation of clustering quality based on the above raw data tables are presented as in the below two subsections.

## 1.1. Data on performance evaluation

This subsection presents the performance evaluation of ECA* with its competitors over 32 datasets. According to [2,3], the clustering results are valided using objective function and cluster quality measures. Cluster quality measures require ground truth labels, whereas objective function measures do not necessitate ground truth labels.

Three cluster quality measures and three objective function measures are used to assess ECA* against its corresponding techniques. We deploy approximation ratio (e-ratio), mean squared error (MSE) and sum of squared error (SSE) as cluster quality measures. Because of three factors, these three cluster quality measures are effective and adequate for evaluating clustering results: (1) SSE is used to measure the average squared error and the average squared difference between the real and expected values; (2) MSE is used to measure the average squared error and the average squared difference between the real and expected values; and (3) Using the e-ratio to equate the results to theoretical results for approximation techniques.

For the objective functions, we use normalised mutual knowledge (NMI), centroid similarity index (CSI) and centroid index (CI). Because of three factors, these three objective function measures are effective and adequate for evaluating clustering results: (1) Using CI to see how many clusters lack centroids or have more than one centroid; (2) To provide a more precise result at the cost of intuitively interpreting the loss of meaning, use CSI to measure point-level differences in the matched clusters and calculate the proportional number of similar points between the matched clusters.; and (3) Using NMI to gain a better understanding of the point-level variations between the matched clusters. The equalions for calculating the aforementioned objective function and cluster quality measures are discussed in [2,3].

**Table 1**
Objective function and cluster quality measures for KM for 30 average run.

| Benchmark datasets | Class qualities | | | Objective functions | | |
| --- | --- | --- | --- | --- | --- | --- |
| | CI | CSI | NMI | SSE | nMSE | $\varepsilon$- ratio |
| S1 | 0 | 0.681 | 1 | 1.22E+13 | 1.22E+09 | 1.21709E+16 |
| S2 | 0 | 0.4935 | 1 | 1.85E+13 | 1.85E+09 | 1.85373E+16 |
| S3 | 0 | 0.5175 | 1 | 2.28E+13 | 2.28E+09 | 2.28352E+16 |
| S4 | 0 | 0.6405 | 1 | 1.69E+13 | 1.69E+09 | 1.68505E+16 |
| A1 | 0 | 0.613 | 1 | 1.87E+10 | 3.12E+06 | 1.87357E+13 |
| A2 | 0 | 0.584 | 1 | 3.80E+10 | 3.62E+06 | 3.79691E+13 |
| A3 | 0 | 0.8295 | 1 | 5.23E+10 | 3.48E+06 | 5.22512E+13 |
| Birch1 | 0 | 0.645 | 0.6144 | 2.10E+11 | 1.05E+06 | 2.10E+14 |
| Un-balance | 0 | 0.503 | 1 | 2.19E+12 | 1.68E+08 | 2.19E+15 |
| Aggregation | 0 | 1 | 0.967 | 1.18E+04 | 7.49E+00 | 1.18E+07 |
| Compound | 0 | 1 | 1 | 5.58E+03 | 6.99E+00 | 5.58E+06 |
| Path-based | 0 | 1 | 0.923 | 8.96E+03 | 1.49E+01 | 8.96E+06 |
| D31 | 0 | 1 | 0.95 | 5.25E+04 | 8.47E+00 | 5.25E+07 |
| R15 | 0 | 1 | 1 | 1.64E+02 | 1.36E-01 | 1.64E+05 |
| Jain | 0 | 1 | 1 | 2.38E+04 | 3.19E+01 | 2.38E+07 |
| Flame | 0 | 1 | 0.866 | 3.19E+03 | 6.65E+00 | 3.19E+06 |
| Dim-32 | 0 | 0.886 | 0.763289668 | 3.79E+07 | 1.16E+03 | 3.79E+10 |
| Dim-64 | 0 | 0.884 | 0.660993808 | 6.23E+07 | 9.51E+02 | 6.23E+10 |
| Dim-128 | 0 | 0.78 | 0.537426286 | 2.45E+09 | 1.87E+04 | 2.45E+12 |
| Dim-256 | 0 | 0.689 | 0.423074312 | 1.76E+08 | 6.73E+02 | 1.76E+11 |
| Dim-512 | 0 | 0.574 | 0.3176381 | 2.62E+08 | 4.99E+02 | 2.62E+11 |
| Dim-1024 | 0 | 0.662 | 0.204291828 | 7.47E+08 | 7.13E+02 | 7.47E+11 |
| G2-16-10 | 0 | 1 | 0.715302092 | 3.28E+06 | 1.00E+02 | 3.28E+09 |
| G2-16-30 | 0 | 1 | 0.613189239 | 2.92E+07 | 8.91E+02 | 2.92E+10 |
| G2-16-60 | 0 | 0.997 | 0.53243349 | 1.17E+08 | 3.56E+03 | 1.17E+11 |
| G2-16-80 | 0 | 0.995 | 0.493047228 | 2.09E+08 | 6.38E+03 | 2.09E+11 |
| G2-16-100 | 0 | 0.996 | 0.527549222 | 4.87E+08 | 1.49E+04 | 4.87E+11 |
| G2-1024-10 | 0 | 0.999 | 0.687949571 | 2.10E+08 | 1.00E+02 | 2.10E+11 |
| G2-1024-30 | 0 | 0.998 | 0.583192724 | 1.88E+09 | 8.98E+02 | 1.88E+12 |
| G2-1024-60 | 0 | 0.997 | 0.512880296 | 7.54E+09 | 3.59E+03 | 7.54E+12 |
| G2-1024-80 | 0 | 0.996 | 0.490720778 | 1.34E+10 | 6.39E+03 | 1.34E+13 |
| G2-1024-100 | 0 | 0.996 | 0.472544466 | 4.25E+10 | 2.03E+04 | 4.25E+13 |

At first glance, the results of objective function and cluster quality measures for ECA* for 30 avrage run are published in [2]. It shows the implementation of ECA* across 32 datasets. Cluster qualities (NMI, CSI and CI) are used as evaluation criteria, as are objective functions (e-ratio,nMSE and SSE). In the vast majority of situations, ECA* is the only successful algorithm among its competitors.

In regard with KM, the results of objective function and cluster quality measures for KM for 30 average run are presented in Table 1. For CI measure, KM determines the correct cluster number for all the 32 datasets, whereas it does not perfrom well in determing the cluster centroid index accurately on 22 datasets. Meanwhile, for the noramlised mutual information, KM has perfect correlation for 11 datasets between its results and ground truth results. On the contrary, KM is a loser in all scenarios for the mean values of the objective functions criteria (e-ratio, nMSE and SSE) for 32 varied datasets.

For KM++, the results of objective function and cluster quality measures for KM for 30 average run are summarised in Table 2. Similar to KM, KM++ determines the correct cluster number for all the 32 datasets, whereas it could not specifically determine the cluster centroid index on 31 datasets. At the meantile, for the noramlised mutual information, KM has perfect relation for 8 datasets between its results and ground truth results. Moreover, KM++ is the winning algorithm in two datasets (G2-1024-100 and A3) for the mean values of the objective function criteria (e-ratio, nMSE and SSE) for 32 datasets.

**Table 2**
Objective function and cluster quality measures for KM++ for 30 average run.

| Benchmark datasets | Class qualities | | | Objective functions | | |
|---|---|---|---|---|---|---|
| | CI | CSI | NMI | SSE | nMSE | $\varepsilon$- ratio |
| S1 | 0 | 0.7897 | 1 | 1.42E+13 | 1.42E+09 | 1.42349E+16 |
| S2 | 0 | 0.9595 | 1 | 1.33E+13 | 1.33E+09 | 1.32812E+16 |
| S3 | 0 | 0.87195 | 1 | 1.89E+13 | 1.89E+09 | 1.89186E+16 |
| S4 | 0 | 0.58025 | 1 | 1.67E+13 | 1.67E+09 | 1.67055E+16 |
| A1 | 0 | 0.954 | 1 | 1.27E+10 | 2.12E+06 | 1.27065E+13 |
| A2 | 0 | 0.9395 | 0.997 | 3.37E+10 | 3.21E+06 | 3.37495E+13 |
| A3 | 0 | 0.982 | 1 | 4.11E+10 | 2.74E+06 | 4.10573E+13 |
| Birch1 | 0 | 0.669 | 0.6712 | 2.10E+10 | 1.05E+05 | 2.10E+13 |
| Un-balance | 0 | 0.818 | 1 | 1.17E+12 | 9.03E+07 | 1.17E+15 |
| Aggregation | 0 | 7.8215 | 0.823880845 | 1.38E+04 | 8.73E+00 | 1.38E+07 |
| Compound | 0 | 6.1285 | 0.767 | 5.71E+03 | 7.16E+00 | 5.71E+06 |
| Path-based | 0 | 0.3445 | 0.826 | 8.96E+03 | 1.49E+01 | 8.96E+06 |
| D31 | 0 | 0.7665 | 0.7 | 5.18E+03 | 8.35E-01 | 5.18E+06 |
| R15 | 0 | 1 | 1 | 1.09E+02 | 9.05E-02 | 1.09E+05 |
| Jain | 0 | 0.205 | 0.866 | 2.38E+04 | 3.19E+01 | 2.38E+07 |
| Flame | 0 | 0.225 | 0.666 | 3.12E+03 | 6.51E+00 | 3.12E+06 |
| Dim-32 | 0 | 0.997 | 0.761058921 | 2.33E+05 | 7.10E+00 | 2.33E+08 |
| Dim-64 | 0 | 0.729 | 0.663447646 | 6.52E+07 | 9.94E+02 | 6.52E+10 |
| Dim-128 | 0 | 0.999 | 0.551829882 | 5.68E+07 | 4.34E+02 | 5.68E+10 |
| Dim-256 | 0 | 0.998 | 0.446266703 | 4.20E+08 | 1.60E+03 | 4.20E+11 |
| Dim-512 | 0 | 0.88 | 0.312428279 | 2.77E+08 | 5.28E+02 | 2.77E+11 |
| Dim-1024 | 0 | 0.934 | 0.249424559 | 2.75E+05 | 2.63E-01 | 2.75E+08 |
| G2-16-10 | 0 | 0.998 | 0.715302092 | 3.28E+06 | 1.00E+02 | 3.28E+09 |
| G2-16-30 | 0 | 0.998 | 0.613189239 | 2.26E+08 | 6.90E+03 | 2.26E+11 |
| G2-16-60 | 0 | 0.997 | 0.53243349 | 1.17E+08 | 3.56E+03 | 1.17E+11 |
| G2-16-80 | 0 | 0.995 | 0.488677777 | 2.09E+08 | 6.38E+03 | 2.09E+11 |
| G2-16-100 | 0 | 0.996 | 0.534322501 | 3.27E+08 | 9.97E+03 | 3.27E+11 |
| G2-1024-10 | 0 | 0.99 | 0.687949571 | 2.10E+08 | 1.00E+02 | 2.10E+11 |
| G2-1024-30 | 0 | 0.998 | 0.583192724 | 1.88E+09 | 8.98E+02 | 1.88E+12 |
| G2-1024-60 | 0 | 0.997 | 0.512880296 | 7.54E+09 | 3.59E+03 | 7.54E+12 |
| G2-1024-80 | 0 | 0.996 | 0.490720778 | 1.34E+10 | 6.39E+03 | 1.34E+13 |
| G2-1024-100 | 0 | 0.996 | 0.472544466 | 2.09E+10 | 9.98E+03 | 2.09E+13 |

For EM, the results of objective function and cluster quality measures for EM for 30 average run are summarised in Table 3. Like to KM and KM++, EM determines the correct cluser number for all the 32 datasets, whereas it could not specifically determine the cluster centroid index on 31 datasets. In the meantime, for the noramlised mutual information, EM has perfect relation for 8 datasets between its results and ground truth results. It is important to mention that EM is the winning algorithm for S1 and G2-1024-100 datasets for the mean values of the objective function criteria (e-ratio, nMSE and SSE) for 32 datasets.

The results of objective function and cluster quality measures for LVQ for 30 average run are summarised in Table 4. Like the algorithms presented previously, LVQ finds the correct number of clusters for all the 32 datasets, whereas it could not specifically determine the cluster centroid index on 31 datasets. At the meantime, for the noramlised mutual information, LVQ has perfect relation for 8 datasets between its results and ground truth results. It is worth mentioning that LVQ is the winning algorithm for G2-16-100, Dim-512 and G2-1024-100 datasets for the mean values of the objective function criteria (e-ratio, nMSE and SSE) for 32 datasets.

Moreover, the results of objective function and cluster quality measures for GENCLUST++ for 30 average run are summarised in Table 5. Similar to its counterpart algorithms, GENCLUST++ determines the correct cluster number for all the 32 datasets, whereas it could specifically determine the cluster centroid index on only one dataset. At the meantime, for the noramlised mutual information, LVQ does not have perfect relation between its results and ground truth results on any dataset. For the mean values of the internal evaluation criteria (e-ratio nMSE and SSE) for 32 datasets, GENCLUST++ is the second winning algorithm, after ECA∗, that has strong performance in four datasets (G2-16-80, S2, Aggregation and Flmae).

**Table 3**
Objective function and cluster quality measures for EM for 30 average run.

| Benchmark datasets | Class qualities | | | Objective functions | | |
|---|---|---|---|---|---|---|
| | CI | CSI | NMI | SSE | nMSE | $\varepsilon$- ratio |
| S1 | 0 | 0.9865 | 1 | 9.09E+12 | 9.09E+08 | 9.08509E+15 |
| S2 | 0 | 0.9795 | 1 | 1.39E+13 | 1.39E+09 | 1.38939E+16 |
| S3 | 0 | 0.963 | 1 | 1.19E+12 | 1.19E+08 | 1.19251E+15 |
| S4 | 0 | 0.9005 | 1 | 2.27E+13 | 2.27E+09 | 2.2668E+16 |
| A1 | 0 | 0.9825 | 1 | 1.58E+10 | 2.63E+06 | 1.57765E+13 |
| A2 | 0 | 0.9425 | 1 | 3.27E+10 | 3.11E+06 | 3.26648E+13 |
| A3 | 0 | 0.9935 | 1 | 4.51E+10 | 3.01E+06 | 4.50839E+13 |
| Birch1 | 0 | 0.717 | 0.7191 | 1.75E+10 | 8.76E+04 | 1.75E+13 |
| Un-balance | 0 | 0.921 | 1 | 1.33E+12 | 1.02E+08 | 1.33E+15 |
| Aggregation | 0 | 0.978 | 0.853 | 1.38E+04 | 8.73E+00 | 13765028.95 |
| Compound | 0 | 1 | 0.876 | 5.63E+03 | 7.06E+00 | 5.63E+06 |
| Path-based | 0 | 0.994 | 0.852 | 9.63E+03 | 1.60E+01 | 9.63E+06 |
| D31 | 0 | 0.9945 | 0.899 | 4.15E+03 | 6.69E-01 | 4.15E+06 |
| R15 | 0 | 1 | 0.831 | 1.70E+02 | 1.42E-01 | 1.70E+05 |
| Jain | 0 | 1 | 0.866 | 2.53E+04 | 3.39E+01 | 2.53E+07 |
| Flame | 0 | 0.9195 | 0.667 | 3.33E+03 | 6.94E+00 | 3.33E+06 |
| Dim-32 | 0 | 0.665 | 0.763289668 | 7.56E+06 | 2.31E+02 | 7.56E+09 |
| Dim-64 | 0 | 0.602 | 0.660993808 | 5.58E+07 | 8.52E+02 | 5.58E+10 |
| Dim-128 | 0 | 0.78 | 0.537426286 | 1.09E+08 | 8.33E+02 | 1.09E+11 |
| Dim-256 | 0 | 0.689 | 0.423074312 | 4.81E+07 | 1.83E+02 | 4.81E+10 |
| Dim-512 | 0 | 0.507 | 0.3176381 | 6.57E+08 | 1.25E+03 | 6.57E+11 |
| Dim-1024 | 0 | 0.662 | 0.204291828 | 5.60E+08 | 5.34E+02 | 5.60E+11 |
| G2-16-10 | 0 | 0.999 | 0.715302092 | 1.67E+08 | 5.09E+03 | 1.67E+11 |
| G2-16-30 | 0 | 0.998 | 0.613189239 | 2.92E+07 | 8.91E+02 | 2.92E+10 |
| G2-16-60 | 0 | 0.997 | 0.53243349 | 1.17E+08 | 3.56E+03 | 1.17E+11 |
| G2-16-80 | 0 | 0.995 | 0.490547526 | 2.09E+08 | 9.98E+01 | 2.09E+11 |
| G2-16-100 | 0 | 0.997 | 0.531515249 | 3.27E+08 | 1.56E+02 | 3.27E+11 |
| G2-1024-10 | 0 | 0.999 | 0.687949571 | 2.10E+08 | 1.00E+02 | 2.10E+11 |
| G2-1024-30 | 0 | 0.998 | 0.583192724 | 1.88E+09 | 8.98E+02 | 1.88E+12 |
| G2-1024-60 | 0 | 0.997 | 0.512880296 | 7.54E+09 | 3.59E+03 | 7.54E+12 |
| G2-1024-80 | 0 | 0.996 | 0.490720778 | 1.34E+10 | 6.39E+03 | 1.34E+13 |
| G2-1024-100 | 0 | 0.996 | 0.472544466 | 2.09E+10 | 9.98E+03 | 2.09E+13 |

### 1.2. Data of performance rating framework

In this subsection, the performance rating framework's data is presented based on five dataset features: Cluster shape; Well-structured clusters (structure); Cluster dimensionality; The number of clusters; and Cluster overlap. The below tables are related to the total average performance rate of ECA∗ and its competitors. For both external and internal validation measures presented in the below tables, The best efficient algorithm to a current factor is rate 1, while the worst efficient algorithm to a current factor is rate 6.

Table 6 presents the total average performance rate of ECA∗ for 32 datasets on the basis of five dataset properies. ECA∗ performs well in all the cluster measures towards the five different dataset features. Exceptionally, ECA∗ has the second rank according to NMI for Number of classes and Class shape. Meanwhile, it has the second rank for Number of classes according to nMSE.

Table 7 shows the total average performance rate of KM for 32 datasets on the basis of five dataset properies. Compared to ECA∗, KM is not performing well except for CI measure. Meanwhile, it has the first rank for Number of clusters and Class shape in accordance with NMI measure. It also ranks the first for Class shape in respect with CSI.

**Table 4**
Objective function and cluster quality measures for LVQ for 30 average run.

| Benchmark datasets | Class qualities | | | Objective functions | | |
|---|---|---|---|---|---|---|
| | CI | CSI | NMI | SSE | nMSE | $\varepsilon$- ratio |
| S1 | 0 | 0.6975 | 1 | 3.11E+13 | 3.11E+09 | 3.10677E+16 |
| S2 | 0 | 0.704 | 1 | 4.75E+13 | 4.75E+09 | 4.75115E+16 |
| S3 | 0 | 0.7585 | 1 | 4.64E+13 | 4.64E+09 | 4.63986E+16 |
| S4 | 0 | 0.8245 | 1 | 3.54E+13 | 3.54E+09 | 3.53919E+16 |
| A1 | 0 | 0.981 | 1 | 5.99E+10 | 9.99E+06 | 5.99487E+13 |
| A2 | 0 | 0.9635 | 1 | 8.13E+10 | 7.74E+06 | 8.13209E+13 |
| A3 | 0 | 0.8855 | 1 | 1.14E+11 | 7.62E+06 | 1.1428E+14 |
| Birch1 | 0 | 0.278 | 0.7615 | 4.20E+11 | 2.10E+06 | 4.20E+14 |
| Un-balance | 0 | 0.9515 | 0.935 | 5.46E+12 | 4.20E+08 | 5.46E+15 |
| Aggregation | 0 | 0.938 | 0.737 | 6.07E+04 | 3.85E+01 | 6.07E+07 |
| Compound | 0 | 0.973 | 0.832 | 1.46E+04 | 1.84E+01 | 1.46E+07 |
| Path-based | 0 | 0.9545 | 0.862 | 1.72E+04 | 2.86E+01 | 1.72E+07 |
| D31 | 0 | 0.9715 | 0.66 | 1.03E+04 | 1.66E+00 | 1.03E+07 |
| R15 | 0 | 1 | 0.821 | 2.10E+02 | 1.75E-01 | 2.10E+05 |
| Jain | 0 | 0.9435 | 1 | 4.31E+04 | 5.77E+01 | 4.31E+07 |
| Flame | 0 | 0.975 | 0.866 | 3.35E+03 | 6.98E+00 | 3.35E+06 |
| Dim-32 | 0 | 0.997 | 0.723360532 | 2.36E+05 | 7.20E+00 | 2.36E+08 |
| Dim-64 | 0 | 0.937 | 0.623579052 | 2.17E+05 | 3.31E+00 | 2.17E+08 |
| Dim-128 | 0 | 0.997 | 0.53346642 | 2.69E+05 | 2.05E+00 | 2.69E+08 |
| Dim-256 | 0 | 0.533 | 0.455016631 | 2.43E+05 | 9.28E-01 | 2.43E+08 |
| Dim-512 | 0 | 0.506 | 0.319000906 | 2.97E+05 | 5.66E-01 | 2.97E+08 |
| Dim-1024 | 0 | 0.507 | 0.309884044 | 1.24E+09 | 1.19E+03 | 1.24E+12 |
| G2-16-10 | 0 | 0.999 | 0.715302092 | 3.28E+06 | 1.00E+02 | 3.28E+09 |
| G2-16-30 | 0 | 0.998 | 0.613189239 | 2.92E+07 | 8.91E+02 | 2.92E+10 |
| G2-16-60 | 0 | 0.933 | 0.406866736 | 1.91E+08 | 5.83E+03 | 1.91E+11 |
| G2-16-80 | 0 | 0.966 | 0.477256591 | 2.50E+08 | 1.19E+02 | 2.50E+11 |
| G2-16-100 | 0 | 0.908 | 0.433097594 | 4.02E+08 | 1.92E+02 | 4.02E+11 |
| G2-1024-10 | 0 | 0.999 | 0.687949571 | 2.10E+08 | 1.00E+02 | 2.10E+11 |
| G2-1024-30 | 0 | 0.998 | 0.583192724 | 1.88E+09 | 8.98E+02 | 1.88E+12 |
| G2-1024-60 | 0 | 0.997 | 0.512880296 | 7.54E+09 | 3.59E+03 | 7.54E+12 |
| G2-1024-80 | 0 | 0.996 | 0.490720778 | 1.34E+10 | 6.39E+03 | 1.34E+13 |
| G2-1024-100 | 0 | 0.996 | 0.472544466 | 2.09E+10 | 9.98E+03 | 2.09E+13 |

Table 8 demonstrates the total average performance rate of KM++ for 32 datasets on the basis of five dataset properies. KM++ is performing well in all dataset features according to CI measure. Similarly, it ranks the first for Number of classes in respect with nMSE.

Table 9 is about the total average performance rate of EM for 32 datasets on the basis of five dataset properies. Similar to previous algoritms, EM is performing well for all dataset features based on CI measure. Additionally, it has the first rank for Class overlap refering to CIS criterion.

Table 10 illustrates rhe total average performance rate of LVQ for 32 datasets on the basis of five dataset properies. LVQ is only performing well for all dataset features based on CI measure, but it is not well performed for the dataset features based on the other valudation measures.

Table 11 is the total average performance rate of GENCLUST++ for 32 datasets on the basis of five dataset properies. GENCLUST++ is solely perfoming well in Class structure dataset feature for NMI.

Overall, Table 12 articulates the grand total average performance rate of ECA*, GENCLUST++, LVQ, EM, KM++ and KM for 32 datasets on the basis of five dataset properies. In this eable, the value of one shows the best performing algorithm for a specified dataset properties, 6 signifies the least performing algorithms for a corresponding dataset feature. Generally, the ordering rate (from the best performing to the least performing one) of the algorithms on the five dataset properties is: (1) ECA*; (2) EM; (3) KM++; (4) KM; (5) LVQ; (6) GENCLUST++.

**Table 5**
Objective function and cluster quality measures for GENCLUST++ for 30 average run.

| Benchmark datasets | Class qualities | | | Objective functions | | |
|---|---|---|---|---|---|---|
| | CI | CSI | NMI | SSE | nMSE | $\varepsilon$- ratio |
| S1 | 0 | 0.807 | 0.972 | 2.14E+13 | 2.14E+09 | 2.13658E+16 |
| S2 | 0 | 0.9165 | 0.994 | 1.32E+13 | 1.32E+09 | 1.32093E+16 |
| S3 | 0 | 0.939 | 0.993 | 1.99E+13 | 1.99E+09 | 1.98653E+16 |
| S4 | 0 | 0.8315 | 0.972 | 2.12E+13 | 2.12E+09 | 2.11831E+16 |
| A1 | 0 | 0.947 | 0.865 | 4.89E+10 | 8.15E+06 | 4.88863E+13 |
| A2 | 0 | 0.97 | 0.97 | 3.36E+10 | 3.20E+06 | 3.35807E+13 |
| A3 | 0 | 0.954 | 0.988 | 5.58E+10 | 3.72E+06 | 5.58414E+13 |
| Birch1 | 22 | 0.23 | 0.7611 | 8.41E+09 | 4.20E+04 | 8.41E+12 |
| Un-balance | 0 | 0.8005 | 0.87 | 5.97E+12 | 4.59E+08 | 5.97E+15 |
| Aggregation | 0 | 0.963 | 0.796 | 9.02E+03 | 5.73E+00 | 9.02E+06 |
| Compound | 1 | 0.9395 | 0.654 | 8.54E+03 | 1.07E+01 | 8.54E+06 |
| Path-based | 0 | 0.887 | 0.39 | 4.62E+03 | 7.70E+00 | 4.62E+06 |
| D31 | 0 | 0.958 | 0.704 | 5.15E+05 | 8.31E+01 | 5.15E+08 |
| R15 | 0 | 0.955 | 0.723 | 3.24E+02 | 2.70E-01 | 3.24E+05 |
| Jain | 0 | 0.9435 | 0.397 | 2.72E+03 | 3.64E+00 | 2.72E+06 |
| Flame | 0 | 0.975 | 0.798 | 1.24E+03 | 2.58E+00 | 1.24E+06 |
| Dim-32 | 3 | 0.415 | 0.7543 | 2.32E+08 | 7.09E+03 | 2.32E+11 |
| Dim-64 | 0 | 0.548 | 0.675 | 3.86E+08 | 5.89E+03 | 3.86E+11 |
| Dim-128 | 3 | 0.543 | 0.491 | 1.79E+07 | 1.36E+02 | 1.79E+10 |
| Dim-256 | 4 | 0.428 | 0.389 | 9.50E+07 | 3.62E+02 | 9.50E+10 |
| Dim-512 | 9 | 0.367 | 0.296 | 5.91E+08 | 1.13E+03 | 5.91E+11 |
| Dim-1024 | 5 | 0.454 | 0.198 | 3.11E+05 | 2.96E-01 | 3.11E+05 |
| G2-16-10 | 0 | 1 | 0.709112968 | 3.28E+06 | 1.00E+02 | 3.28E+09 |
| G2-16-30 | 0 | 0.998 | 0.599823022 | 2.92E+07 | 8.92E+02 | 2.92E+10 |
| G2-16-60 | 0 | 0.996 | 0.499045271 | 1.17E+08 | 3.56E+03 | 1.17E+11 |
| G2-16-80 | 0 | 0.995 | 0.494278534 | 2.10E+08 | 1.00E+02 | 2.10E+11 |
| G2-16-100 | 0 | 0.951 | 0.484199364 | 5.41E+08 | 2.58E+02 | 5.41E+11 |
| G2-1024-10 | 0 | 0.953 | 0.548988802 | 1.33E+10 | 6.36E+03 | 1.33E+13 |
| G2-1024-30 | 0 | 0.953 | 0.550963578 | 1.50E+10 | 7.16E+03 | 1.50E+13 |
| G2-1024-60 | 0 | 0.952 | 0.5029802 | 2.07E+10 | 9.85E+03 | 2.07E+13 |
| G2-1024-80 | 1 | 0.969 | 0.502981 | 1.34E+10 | 6.39E+03 | 1.34E+13 |
| G2-1024-100 | 0 | 0.849 | 0.336583377 | 2.62E+10 | 1.25E+04 | 2.62E+13 |

**Table 6**
The total average performance rate of ECA* for 32 datasets on the basis of five dataset properies.

| Cluster characteristics | Cluster measure qualities | | | Cluster objective functions | | |
|---|---|---|---|---|---|---|
| | CI | CSI | NMI | SSE | nMSE | $\varepsilon$- ratio |
| **Class overlap** | 1 | 1 | 1 | 1 | 1 | 1 |
| **Number of classes** | 1 | 1 | 2 | 1 | 2 | 1 |
| **Class dimensionality** | 1 | 1 | 1 | 1 | 1 | 1 |
| **class structure** | 1 | 1 | 1 | 1 | 1 | 1 |
| **Class shape** | 1 | 1 | 2 | 1 | 1 | 1 |

**Table 7**
The total average performance rate of KM for 32 datasets on the basis of five dataset properies.

| Cluster characteristics | Cluster measure qualities | | | Cluster objective functions | | |
|---|---|---|---|---|---|---|
| | CI | CSI | NMI | SSE | nMSE | $\varepsilon$- ratio |
| **Class overlap** | 1 | 5 | 2 | 4 | 4 | 4 |
| **Number of classes** | 1 | 5 | 1 | 4 | 4 | 4 |
| **Class dimensionality** | 1 | 4 | 3 | 4 | 4 | 6 |
| **class structure** | 1 | 4 | 2 | 5 | 5 | 5 |
| **Class shape** | 1 | 1 | 1 | 6 | 4 | 4 |

**Table 8**
The total average performance rate of KM++ for 32 datasets on the basis of five dataset properies.

| Cluster characteristics | Cluster measure qualities | | | Cluster objective functions | | |
|---|---|---|---|---|---|---|
| | CI | CSI | NMI | SSE | nMSE | $\varepsilon$- ratio |
| **Class overlap** | 1 | 2 | 2 | 3 | 3 | 3 |
| **Number of classes** | 1 | 6 | 4 | 2 | 1 | 2 |
| **Class dimensionality** | 1 | 2 | 2 | 2 | 2 | 4 |
| **class structure** | 1 | 3 | 5 | 4 | 4 | 4 |
| **Class shape** | 1 | 5 | 5 | 2 | 2 | 2 |

**Table 9**
The total average performance rate of EM for 32 datasets on the basis of five dataset properies.

| Cluster characteristics | Cluster measure qualities | | | Cluster objective functions | | |
|---|---|---|---|---|---|---|
| | CI | CSI | NMI | SSE | nMSE | $\varepsilon$- ratio |
| **Class overlap** | 1 | 1 | 2 | 2 | 2 | 2 |
| **Number of classes** | 1 | 2 | 5 | 3 | 3 | 3 |
| **Class dimensionality** | 1 | 5 | 3 | 3 | 3 | 5 |
| **class structure** | 1 | 2 | 4 | 3 | 3 | 3 |
| **Class shape** | 1 | 2 | 3 | 3 | 3 | 3 |

**Table 10**
The total average performance rate of LVQ for 32 datasets on the basis of five dataset properies.

| Cluster characteristics | Cluster measure qualities | | | Cluster objective functions | | |
|---|---|---|---|---|---|---|
| | CI | CSI | NMI | SSE | nMSE | $\varepsilon$- ratio |
| **Class overlap** | 1 | 4 | 3 | 6 | 6 | 6 |
| **Number of classes** | 1 | 3 | 3 | 6 | 6 | 6 |
| **Class dimensionality** | 1 | 3 | 4 | 5 | 5 | 2 |
| **class structure** | 1 | 5 | 3 | 6 | 6 | 6 |
| **Class shape** | 1 | 3 | 4 | 4 | 6 | 5 |

**Table 11**
The total average performance rate of GENCLUST++ for 32 datasets on the basis of five dataset properies.

| Cluster characteristics | Cluster measure qualities | | | Cluster objective functions | | |
|---|---|---|---|---|---|---|
| | CI | CSI | NMI | SSE | nMSE | $\varepsilon$- ratio |
| **Class overlap** | 2 | 3 | 4 | 5 | 5 | 5 |
| **Number of classes** | 2 | 4 | 6 | 5 | 5 | 5 |
| **Class dimensionality** | 2 | 6 | 5 | 6 | 6 | 3 |
| **class structure** | 2 | 6 | 1 | 2 | 2 | 2 |
| **Class shape** | 2 | 4 | 6 | 5 | 5 | 6 |

**Table 12**
The grand total average ranking table of ECA* with its competitors for 32 datasets.

| Cluster characteristics | ECA* | KM | KM++ | EM | LVQ | GENCLUST++ |
|---|---|---|---|---|---|---|
| **Class overlap** | 1.33 | 3.17 | 2.67 | 2.83 | 4.17 | 4.50 |
| **Number of classes** | 1.00 | 3.33 | 2.33 | 1.67 | 4.33 | 4.00 |
| **Class dimensionality** | 1.00 | 3.67 | 2.17 | 3.33 | 3.33 | 4.67 |
| **class structure** | 1.00 | 3.67 | 3.50 | 2.67 | 4.50 | 2.50 |
| **Class shape** | 1.17 | 2.83 | 2.83 | 2.50 | 3.83 | 4.67 |
| **Average** | **1.10** | **3.33** | **2.70** | **2.60** | **4.03** | **4.07** |

## 2. Experimental Design, Materials and Methods

ECA* is a ensemble technique primarily used for clustering divergent datasets [2]. The proposed algorithm is mainly based on backtracking search optimisation algorithm (BSA) [4,5], social class ranking, levy flight optimisation, quartiles and percentiles, and Euclidean distance. The datasets shown in this article is gathered from the results of ECA* with its five ompetitive algorithms applied to solve 32 benchmark clustering datasets in three methods as follows:

- Assessing the efficiency of ECA* against its competitor using external and internal measures;
- Analysing the statistical performance of ECA* compare to its correspnding algorithms based on their mean solution (average), the worst solution (worst) and the best solution (best) for execution time, interCluster distance and intraCluster distance;
- It proposes a performance rating framework to explore the performance sensitivity of the algorithms on five various dataset features.

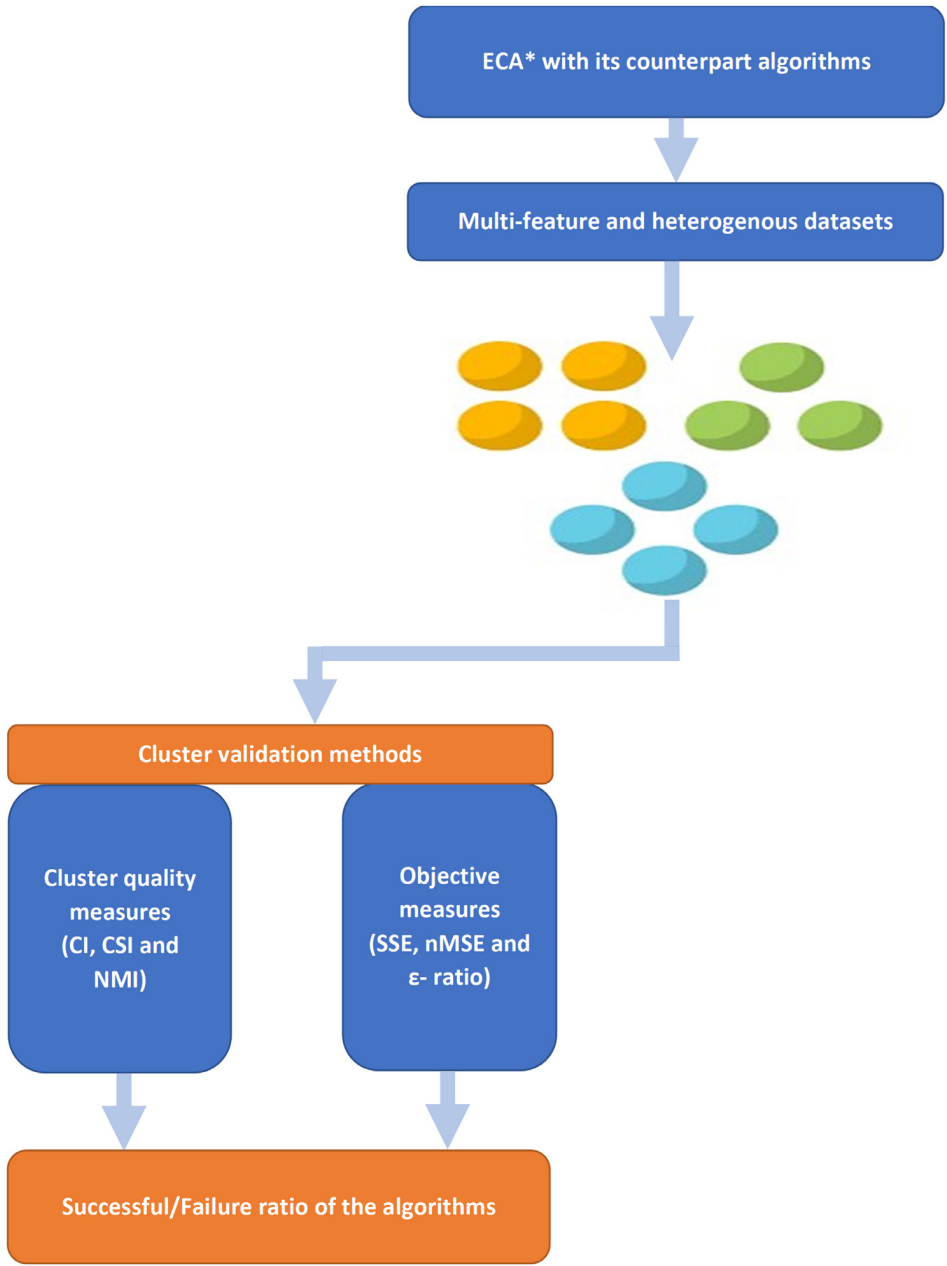

**Fig. 1.** The methodology of validating the clustering results of ECA* compare to its counterpar algorithms.

The comparison is conducted to evaluate the quality of clustering results of ECA* on 32 benchmark datasets in comparison to its competitive algorithms using two clustering evaluation measures. The first measure is cluster quality measures (NMI, CSI and CI), and the second measure is objective function measures ($\varepsilon$- ratio, nMSE and SSE).

The competitive algorithms of ECA* are genetic algorithm for clustering++ (GENCLUST++), learning vector quantisation (LVQ), expectation maximisation (EM), K-means++ (KM++) and K-means (KM). KM divides a set of data points into cluster numbers by summing the square distances between each point and the cluster centroids closest to it [6]. Despite their inaccuracy, K-means' pace and simplicity are appealing to many researchers. However, K-means has several limitations. Fig. 1 depicts the methodology of validating the clustering results of ECA* compare to its counterpar algorithms.

To address the aforementioned flaws, K-means++, a new clustering technique on the basis of KM, was introduced to produce optimum clustering results [7]. In KM++, an efficientapproach is used to select cluster centroids and improve cluster efficiency as a result. Lately, this study [8] developed another technique named EM to maximise the overall probability or likelihood of the data and provide a improved clustering performance. LVQ is another important algorithm for grouping data, developed by [9] to identify patterns in which each output group is labelled by a cluster. Recently, GENCLUST++ was improved by [10] by combining genetic algorithm (GA) with the KM to generate high-quality clusters as a new GA operator re-arrangement. These above-mentioned counterpart algorithms are fully explained in [3].

Since clustering results may differ among various runs, we the aforementioned alorithms 30 times on each dataset to register the clustering results for each execution. Java programming language is used to run ECA*, whereas Weka is utilised to tun its five competitive algorithm per benchmark dataset. We also register the mean results for 30 times run on each mentioned dataset. For the future work, adapting and applying ECA* for real applications is a good possibility., ECA* could be used for application problems [11], e-organisation services [12], ontology learning [15], multi-dimensional database systems [14], and technical and maintenance application [13].

**CRediT Author Statement**

**Bryar A. Hassan:** Conceptualisation, Methodology, Software, Writing - Original Draft, Visualisation, Formal analysis; **Tarik A. Rashid:** Project administration, Investigation, Resources, Data Curation Writing - Review & Editing, Funding acquisition (if applicable), Supervision; **Seyedali Mirjalili:** Data Curation Writing - Review & Editing, Validation, Methodology.

**Funding**

No funding is obtained.

**Declaration of Competing Interest**

There are no conflicts of interest declared by the researchers.

**Acknowledgements**

The researchers would like to express their gratitude to the referees for their insightful comments. Based on their feedback, the technical content of this paper has been greatly improved. Meanwhile, the authors wish to express their gratitude to University of Kurdistan Hewler, Centre for Artificial Intelligence Research and Optimisation and Kurdistan Institution for Strategic Studies and Scientific Research and for their continued support in conducting this research.

**Supplementary Materials**

Supplementary material associated with this article can be found in the online version at doi:10.1016/j.dib.2021.107044.